\documentclass[conference]{IEEEtran}
\IEEEoverridecommandlockouts
\usepackage{cite}
\usepackage{amsmath,amssymb,amsfonts}
\usepackage{graphicx}
\usepackage{textcomp}
\usepackage{xcolor}
\usepackage{algorithm}
\usepackage{algpseudocode}
\usepackage{booktabs}

\usepackage{lipsum}
\def\BibTeX{{\rm B\kern-.05em{\sc i\kern-.025em b}\kern-.08em
    T\kern-.1667em\lower.7ex\hbox{E}\kern-.125emX}}
\begin{document}

\title{A Spatial-Temporal Transformer based Framework For Human Pose Assessment And Correction in Education Scenarios\\
}

\author{\IEEEauthorblockN{Wenyang Hu}
\IEEEauthorblockA{\textit{Fudan-Xiding AI Joint LAB} \\
\textit{Shangshai Xiding AI} \\ \textit{Research Center Co., Ltd}\\
Shanghai, China \\
wlm0909@ieee.org}

\and
\IEEEauthorblockN{Kai Liu}
\IEEEauthorblockA{\textit{R\&D Center} \\
\textit{Shangshai Xiding AI} \\ \textit{Research Center Co., Ltd}\\
Shanghai, China \\
kai.liu@xiding.com.cn}

\and
 
\IEEEauthorblockN{Libin Liu}
\IEEEauthorblockA{\textit{R\&D Center} \\
\textit{Shangshai Xiding AI} \\ \textit{Research Center Co., Ltd}\\
Shanghai, China \\
ronnie.liu@xiding.com.cn} 
\and 
\IEEEauthorblockN{Huiliang Shang*}
\IEEEauthorblockA{\textit{School of Information} \\ \textit{Science and Technology} \\
\textit{Fudan University}\\
Shanghai, China \\
shanghl@fudan.edu.cn}

}

\maketitle

\begin{abstract}
Human pose assessment and correction play a crucial role in applications across various fields, including computer vision, robotics, sports analysis, healthcare, and entertainment. In this paper, we propose a Spatial-Temporal Transformer based Framework (STTF) for human pose assessment and correction in education scenarios such as physical exercises and science experiment. The framework comprising skeletal tracking, pose estimation, posture assessment, and posture correction modules to educate students with professional, quick-to-fix feedback. We also create a pose correction method to provide corrective feedback in the form of visual aids. We test the framework with our own dataset. It comprises (a) new recordings of five exercises, (b) existing recordings found on the internet of the same exercises, and (c) corrective feedback on the recordings by professional athletes and teachers. Results show that our model can effectively measure and comment on the quality of students’ actions. The STTF leverages the power of transformer models to capture spatial and temporal dependencies in human poses, enabling accurate assessment and effective correction of students' movements.
\end{abstract}


\section{Introduction}
Artificial intelligence (AI) has emerged as a disruptive force in many areas include education. There are various education activities requiring pose and movement assessment of student such as physical exercises, instrument operation, engineering practice etc. AI aided pose assessment enables real-time feedback to both teachers and students during educational activities. By accurately analyzing students' poses, teachers can provide immediate feedback on correct posture, body alignment, and movement techniques. This instant feedback allows students to make necessary adjustments, leading to improved posture, motor skills, and overall performance. AI facilitates data analysis on a large scale, allowing educators to gain valuable insights into student performance, engagement, and learning patterns. Educators can identify areas where students may need additional support or intervention. 

Multiple areas in computer vision research have benefited from deeper layers in deep convolutional neural networks (DCNNs). They include image classification~\cite{he2016deep,woo2018cbam}, objection detection~\cite{zhou2019objects,ren2015faster}, human pose
estimation~\cite{wei2016convolutional,wang2020deep}. Sophisticated fitness applications and products now make good use of DCNNs. For example,  Fiture, a smart home fitness mirror, uses DCNN-driven human pose estimation to count exercise repetitions and to recognize exercise type. However, no known effort has been made on using DCNN-driven computer vision research to develop applications that can help students avoid making mistakes.

Posture assessment plays a key role in detecting we during many education scenarios. However, researches~\cite{TDMSC} so far has only investigated static actions in fixed environments. Most images in these datasets were taken while the student was pretending the faulty movement. These incorrect postures could be easily detected by the existing off-the-shelf (OTS) pose estimators~\cite{OpenPose}. However, in real-world scenarios, unexpected situations often occur. These situations include students' hesitation and complicated environment. This is because pose estimation models haven’t been trained to distinguish between the poses and environmental factors such as occlusion and illumination. Therefore, pose estimators fail to detect large fault but instead focus on the subtle deficiencies in these situations. In addition to inefficient error detection, these estimators also lack a criterion for assessing a user’s performance or to providing feedback to correct their actions.
\begin{figure*}
  \centering
  \includegraphics[width=1.0\linewidth]{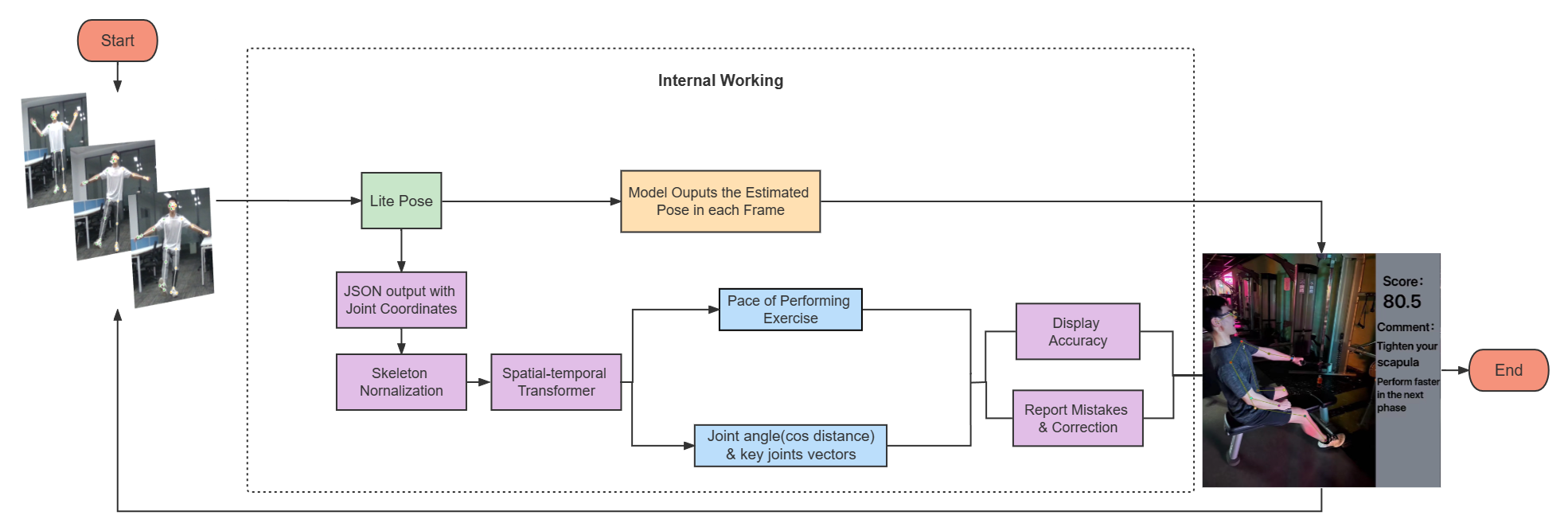}
  \caption{A complete pipeline for our framework}
  \label{fig1}
\end{figure*}

With the help of DCNNs, we create a general framework to evaluate and correct postures of amateurs. Since OTS is largely dependent on the performance of pose estimation, LitePose~\cite{Lite}, a state-of-the-art, lightweight model is an ideal option. Unlike conventional classification networks such as AlexNet~\cite{krizhevsky2012imagenet}, ResNet~\cite{he2016deep}, Litepose integrates high-resolution stream and high-to-low resolution streams together in parallel rather than in series. It allows us to fully exploit the potential of pose estimation, and enables us to focus on the evaluation and correction of an exercise. Furthermore, the limitation of OTS in controlled conditions requires an alternative. Thus, a dataset of real-world raw data is created: our dataset collects the videos mostly from athletes and teachers in time. 

\vspace{0.3cm}

With these preliminaries, we propose our framework (Fig \ref{fig1})for professional pose estimation and corresponding feedback. It outperforms those error-prone pose estimators. The contributions of this paper are summarized as follows:

1.	Novelty in the evaluation of the quality of action: Incorrect postures in exercises would cause other joints to compensate, leading to various kinds of osteoarticular injuries~\cite{taunton2002retrospective}. As a result, we utilize spatial-temporal transformer~\cite{zheng20213d} to assess the mistakes. The model learns to attend to relevant temporal cues and patterns in the actions, allowing it to make predictions about their quality. The self-attention mechanism in the temporal component helps the model focus on important frames or moments within the sequence that contribute to the overall assessment.

2.	Pose correction specialization: In all the videos in the dataset, the reference angles for the targeted joints in each video are annotated. In addition, the range of joint mobility for each joint has been set according to professional assessment~\cite{seow1999study} for normal people. In our algorithm, the model would provide a visualized feedback for the incorrect joints. For skill assessment (SA), there is also a visualized feedback telling the student how to act better. Common skill deficiencies includes: the student does not perform the full range of motion or performs the eccentric phase too fast.

3.	Application to real-world situations: The pattern and representational ability learned by the models in these unlabeled or single object videos can not applied to real world environment, in which all the factors are changeable. As a result, we create a model that is robust to appearance changes in videos while focusing on the posture. We conduct global and local normalization in our in-the-wild dataset to achieve this goal.
\begin{figure*}
  \centering
  \includegraphics[width=.8\linewidth]{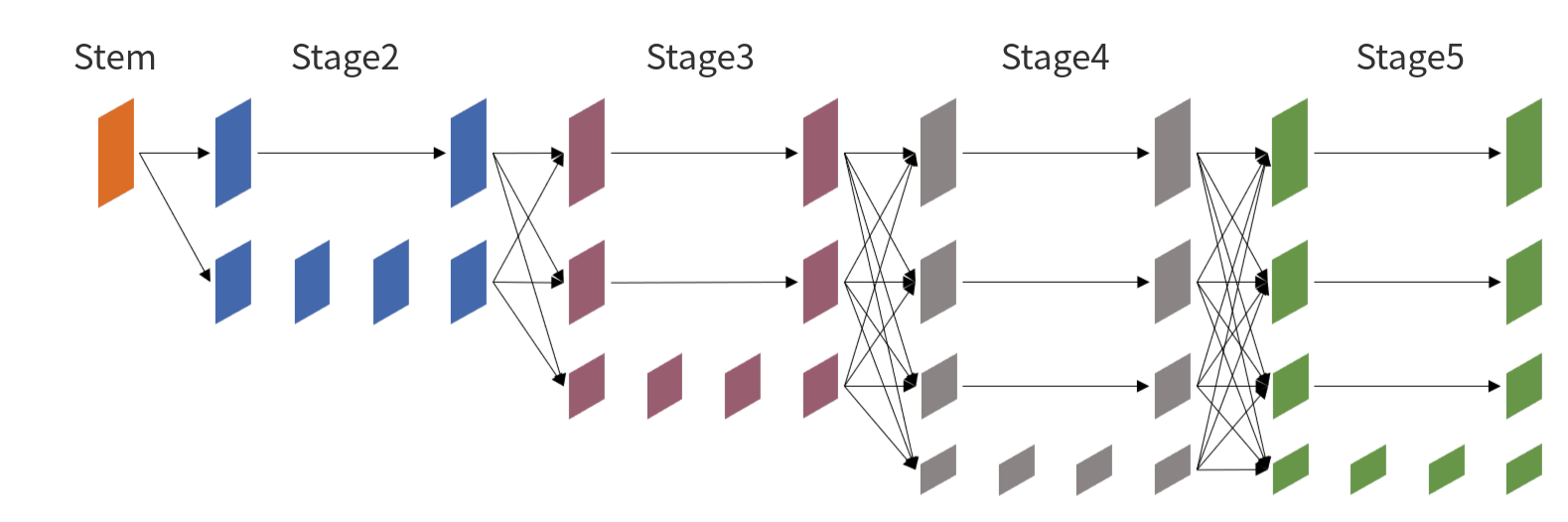}
  \caption{LitePose Architecture}
  \label{fig2}
\end{figure*}
\section{Related work}
\subsection{Human Pose Estimation}
Pose estimation systems were used to produce local interpretations of a single pose before convolutional neural networks (CNN) appeared. CNN, however, is able to learn diverse levels of features according to its depth. Stacked Hourglass Network~\cite{newell2016stacked} combines repeated bottom-up, top-down inference with intermediate supervision to improve the network performance on human pose estimation. It fully exploits CNN's strength in capturing and consolidating information across all scales of the image.  HRNet~\cite{sun2019deep} maintains high-resolution representations through fully convolutional layers~\cite{long2015fully} . It keeps strong position sensitivity by fusing multiple scales of features from multi-resolution streams. Its residual connections~\cite{he2016deep}  remove  redundant layers so that the network will not confuse the right parameters with the wrong ones. Litepose~\cite{Lite}, on the other hand, investigates the effectiveness of single-branch architecture instead. Its experiment conducts gradual shrinking to determine the most appropriate structure for real-time pose estimation models.

\subsection{Action quality assessment}
Action quality assessment(AQA) aims at quantifying the quality of action with distinct criteria to predict a final score. By far, works in AQA have largely limited in areas like physiotherapy~\cite{parmar2019and}, Olympic events~\cite{parmar2019action}. However, our work involved in a larger scale. Therefore, the criteria to quantify AQA scores are different and varied. In general, approaches in AQA can be categorized into two parts: human-pose features based~\cite{Pan_2019_ICCV} and image/video-features based. Video feature based model assesses the whole scene in a coarse way~\cite{parmar2019action} because they only generate a feature representation at video level. However, action quality also depends on the combination of joints. Parmar et al.~\cite{parmar2019and} implement the action quality assessment task for diving using a multi-task learning framework. The framework includes three tasks highly related to action quality assessment: action recognition, video commentary and AQA. Pan et al.~\cite{Pan_2019_ICCV} utilize spatial-temporal relations between joints to obtain the joint kinetic difference and commonality in graph modelling. The evaluation achieves a more comprehensive interpretation through learning the relation graphs for neighbouring joints.

\subsection{Skill assessment}
Gordon et al.~\cite{gordon1995automated} was the first to evaluate the quality of skills through an automated process. It addresses a few issues concerning the application of Automated Video Assessment by tracking the skeleton trajectories. Doughty et al.~\cite{doughty2018s} utilizes CNN with ranking loss as the objective function to evaluate various skills including surgical, drawing, chopstick use and dough rolling. In their subsequent work~\cite{doughty2019pros}, they adopt temporal attention. Liu et al.~\cite{li2019manipulation} proposed a new RNN based spatial attention model to achieve this objective, including skills like baby grasping objects. This method focuses on the real-time visual information of each frame in the video and the high-level relevant knowledge for specific tasks.

\section{Method}
Figure 1 presents an overview of our framework’s pipeline. In our framework, there are three classes of exercise videos: 1. groundtruth videos ($V_{gt}$), 2. correct videos ($V_{c}$), 3. wrong videos ($V_{w}$). We use LitePose to estimate the coordinates of each joint and send them in JSON format. LitePose follows the COCO dataset key point standard, in which 17 key points (nose (1), eyes (2), ears (2), shoulders (2), elbows (2), wrists (2), hip (2), knees (2), and ankles (2)) are detected. Global and local normalizations are applied to these coordinates to alleviate camera distance issues. The final outputs contain frames of different exercises with the same torso size, thereby ensuring the performance of the pose evaluation algorithm. We then apply spatial-temporal transformer to all the videos to let it learn all types of mistakes. These mistakes are organized in two different scales. As long as the model detects these mistakes in the test dataset, frames containing wrong movements will be selected and be transferred into the pose correction module. The pose correction module generates improvement in the form of visual aids.

\subsection{Pose Estimation}
An ideal pose estimator should make accurate predictions for each joint at low computational cost. LitePose emerged as the top candidate in our tests. The model utilizes a single-branch design and incorporates scale fusion at each resolution. In contrast, traditional single-branch deconvolution heads achieve low computation but suffer from low accuracy. LitePose addresses the scale variation problem by conducting a gradual shrinking experiment to assess the impact of removing high-resolution blocks on the accuracy of the final results. When all high-resolution blocks are removed, the resulting architecture closely resembles the hourglass network. LitePose improves performance by adopting multi-scale fusion by connecting only the low-level high-resolution features generated by the deconvolution process in the previous stages with the final prediction layers. The final architecture of LitePose is illustrated in Figure \ref{fig2}

\subsection{Normalization}
We normalize the data by first calculating the pixel distance (torso) between the central points of the two shoulders and the hip. Next, each line connecting two joints is divided by this calculated distance in the extracted skeleton. Therefore, the magnitude of all the lines that form the 2D skeleton is now smaller than 1. Global and local normalization~\cite{sun2017human} is then performed to align the joints with the pose assessment module. The center of the human body is first determined by finding the intersection of the bounding box's two diagonals. The skeleton is then translated into an upright position. We also translate root joints, such as the shoulders, into a same position across different videos. Thus, the model can compare the angular difference between lines connecting middle joints (e.g. elbow) and end joints (e.g. wrist). Specifically, the process follows these steps:

\begin{equation} 
[x^{\rm '}, y^{\rm '}]^{\rm T} = Tr(R^{\rm \alpha}_{x}, R^{\rm \beta}_{y}, d)[x, y, 1]^{\rm T}
\end{equation}
in which $x^{\rm '}, y^{\rm '}$ are the coordinates of the points in the new coordinate system, $d$ is a translation vector and Tr is the translation matrix:
\begin{equation} 
Tr = 
\begin{gathered}
\left[
\begin{array}{ccc}
R^{\alpha}_{x} & 0 \\ 0 & 1
\end{array}
\right] 
\left[
\begin{array}{ccc}
R^{\beta}_{y} & 0 \\ 0 & 1
\end{array}
\right] 
\left[
\begin{array}{ccc}
I_{2} & d \\ 0 & 1
\end{array}
\right] 
\end{gathered}
\end{equation}
$I_{2}$ is a two-dimensional diagonal matrix. Assuming the center point of the hip to be o=[x,y,z], We expressed Tr as: 
\begin{equation} 
Tr = 
\left[
\begin{array}{ccc}
R & d \\ o & 1
\end{array}
\right] 
R = 
\left[
\begin{array}{ccc}
\alpha^{\rm T} & \beta^{\rm T}
\end{array}
\right]^{\rm -1}
d = -o^{\rm T}
\end{equation}

\floatname{algorithm}{Alg}
\begin{algorithm}
\caption{Joint vector using cos similarity}
\label{alg:algorithm-label}
    \begin{algorithmic}[1]
        \Require targeted joints coordinates
		\Ensure Cos distance between adjacent frames
        \For {$iteration=1,2,\ldots$}
        \For {$joint_{tar}=1,2,\ldots,N$}
        \If {$joint_{tar} != iteration$}
        \State Combine each joint with the others for T times
        \State Compute direction vectors $\hat{A}_{1},\ldots,\hat{A}_{T}$
        \EndIf
        \EndFor
        \EndFor
        \label{Alg1}
    \end{algorithmic}
\end{algorithm}

\begin{figure}
  \centering
  \includegraphics[width=.9\columnwidth]{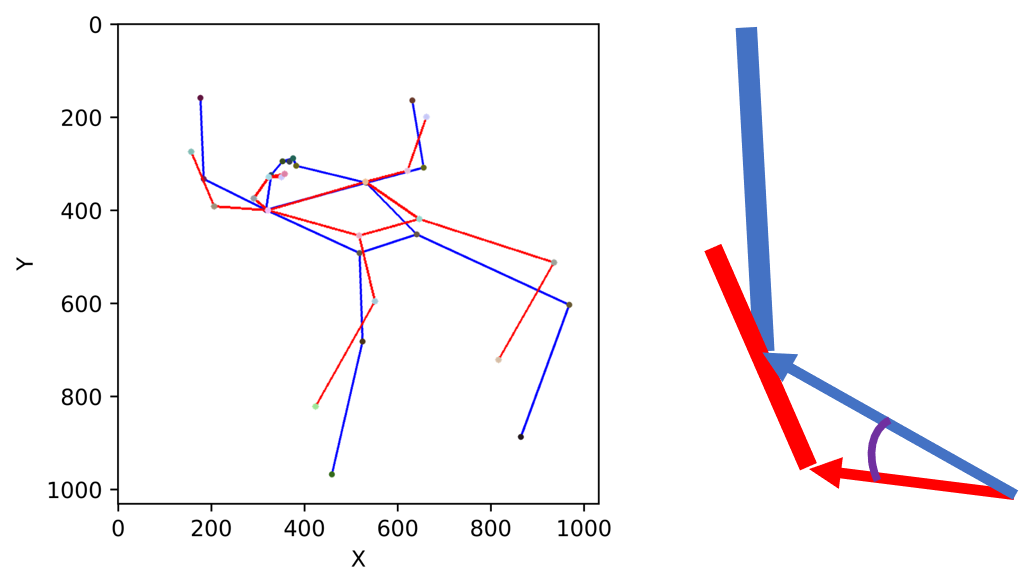}
  \caption{Global and Local Normalization in Bench Press}
  \label{fig3}
\end{figure}

The global and local views of the normalized skeletons are shown in Figure \ref{fig3}.

Normalization for our dataset's labels is also conducted. We determine key joints for assessment by comparing the deviation of each joint’s angles between the first and the last frame. We then construct a N-1 direction vector for each of the N joints. This N*(N-1) vector expresses the relationship between these key joints. The pseudo code for calculating the target joint vector is shown in Alg \ref{Alg1}.

\subsection{Pose Assessment}
We utilize two major assessment criteria as labels for our dataset. They allow the model to provide comprehensive feedback. The first criterion focuses on joint angles, which enables our model to evaluate the alignment and positioning of joints. Additionally, we introduce the criterion of pace, which reflects the student's focus, skill, and potential dangers during exercise. By combining these assessment criteria, our model offers a holistic approach to action quality assessment, promoting optimal performance and minimizing the risk of injuries.

\subsubsection{Spatial-Temporal Transformer}
The spatial-temporal transformer~\cite{zheng20213d} consists of two essential components: the spatial transformer module and the temporal transformer module. The spatial component of the transformer is designed to extract high-dimensional feature embeddings from individual frames. Through the integration of self-attention mechanisms and multiple layers of processing, the spatial transformer effectively aggregates information across all joints, resulting in comprehensive joint sequences of features. This spatial encoding capability is instrumental in identifying fine-grained details and patterns such as interactions between neighboring joints that contribute to action quality. The temporal component of the transformer focuses on modeling the temporal dependencies across multiple frames or images. It takes into account the sequential nature of human motion and examines the progression of actions over time.By flattening the output of the spatial transformer for each frame and concatenating them, the temporal transformer captures the temporal relationships between consecutive frames. This allows the module to assess their adherence to desired criteria and provide a comprehensive representation of the entire sequence of frames. Therefore, this comprehensive understanding of both spatial and temporal aspects empowers the model to evaluate the quality of actions in a holistic manner. 

\subsubsection{Criteria construction}
By examining joint angles, our model assesses the conformity of joint positions to desired benchmarks. This information is instrumental in evaluating aspects such as the proper execution of movements, the extent of joint flexion or extension, and the stability of joints during the action. The analysis of joint angles contributes to a comprehensive understanding of the correctness and effectiveness of the performed actions. For instance, in the context of a deep squat exercise where hip angles, knee angles, and ankle angles are all crucial factors, the calculation of joint angles enables our model to detect deviations or discrepancies in joint positions, thus pinpointing potential mistakes.

Pace also plays a pivotal role in assessing action quality by providing insights into focus, skill, and potential dangers during exercise. Pace not only reflects the one's ability to execute movements effectively but also helps identify areas of improvement and reduce the risk of injuries. For instance, when trainees are doing exercises, they often tend to perform at their maximum speed and thereby overlook core tightening~\cite{kibler2006role}, leading to lower back injuries. When students are doing experiments, they also execute in a higher than average speed, causing potential accidents. To reduce these safety hazards, our model evaluates temporal information of an action, ensuring that an appropriate pace is maintained while upholding proper form and technique. The temporal component of our model enables the analysis of pace-related factors, such as acceleration patterns, time spent in specific positions, and comparisons with reference completion times. 
\begin{figure}
  \centering
  \includegraphics[width=1.0\linewidth]{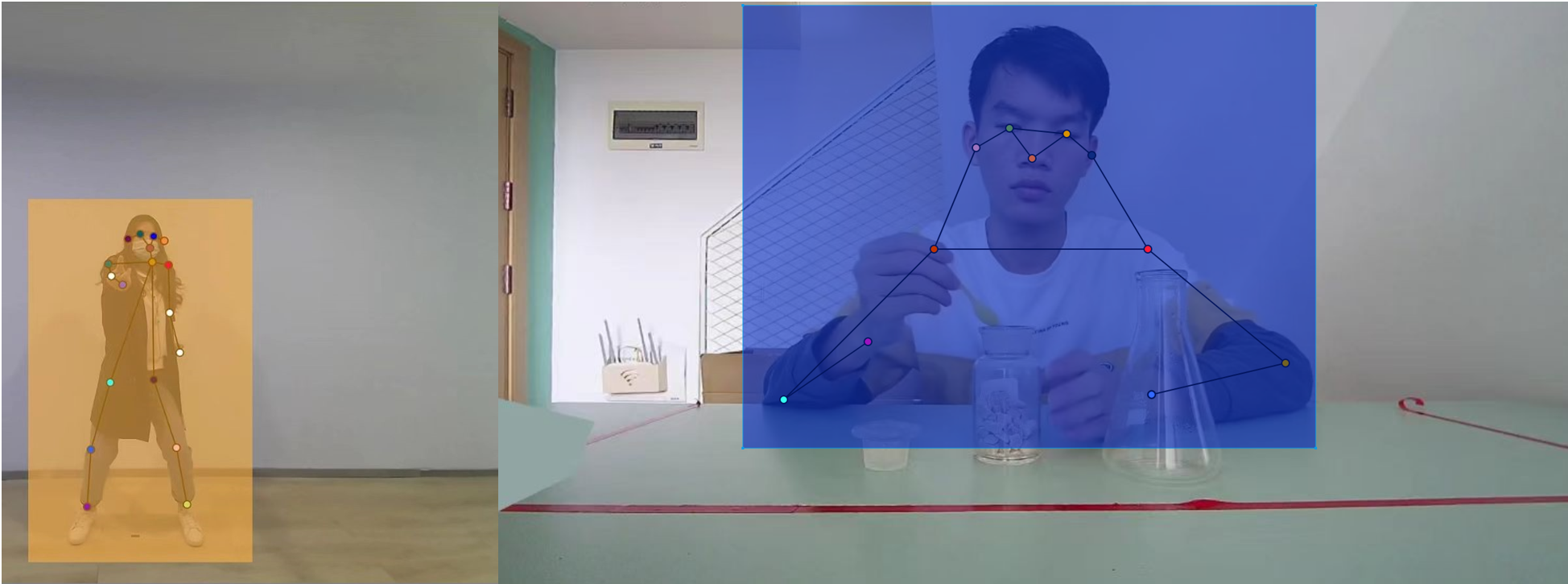}
  \caption{Pose estimation results}
  \label{fig4}
\end{figure}

\begin{figure*}
  \centering
  \includegraphics[width=.9\linewidth]{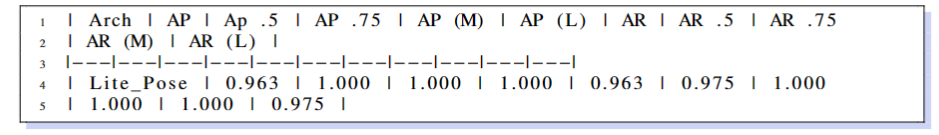}
  \caption{Pose estimation validation results}
  \label{fig5}
\end{figure*}
\subsection{Pose Correction}
Our model employs a visual aid generation technique~\cite{pirsiavash2014assessing} to identify and highlight mistakes in videos. These visual aids are presented in specific key frames. By analyzing input videos with spatial-temporal transformer, mistakes can be localized in certain key frames. When such discrepancies are identified, the model generates a visual aid that demonstrates the correct position for the joints. For example, during a bench press exercise, the primary focus is on the movement of elbow, while wrist and shoulder should remain stationary. If an exercise video demonstrates forearm adduction during the bench press, our model will provide a visual aid illustrating the correct placement of the elbow, which means a right angle between forearm and posterior arm. By leveraging these techniques, our model effectively detects and addresses mistakes in these videos. The visual aids serve as valuable feedback to guide students towards the correct joint positions and angles, promoting proper form and enhancing the quality of actions performed.

\section*{Experiment}
The quantitative and qualitative results of our framework on three different exercises(bench press, pull-up and deep squat) are presented in Table \ref{Tbl1}

\subsection{Pose Estimation}
\subsubsection{COCO Dataset}
The COCO dataset~\cite{lin2014microsoft} comprises over 200,000 labeled images and 250,000 person instances, with annotations for 17 keypoints. It is divided into train/val/test-dev sets, consisting of 57k, 5k, and 20k images, respectively. 

\subsubsection{Implementation Details}
For 2D pose estimation, our experiments utilize the train2017 set for training purposes, while evaluations are conducted on the val2017 and test-dev2017 sets. We employ the standard average precision (AP) as the evaluation metric on the COCO dataset. AP is computed using the Object Keypoint Similarity (OKS) metric, which considers factors such as Euclidean distance between predicted and groundtruth keypoints, keypoint visibility flags (vi), object scale (s), and keypoint-specific constants (ki). For model training, we utilize the Adam optimizer. In the case of LitePose~\cite{Lite}, we adopt the original settings as described in their paper. We also use COCO Annotator to annotate human actions to verify the effectiveness of LitePose on real-world data. The output is presented in Fig \ref{fig4}.

\begin{table}
 \caption{Sample Results of Three Varied Exercises}
  \centering
  \begin{tabular}{llllll}
    \toprule
    Name        & Class & Joint & Pace & Range & Correction\\
    \midrule
    Bench Press(C) & Upper  & 72.2 & 87.5 & 93.8 & Tighten scapula\\
    Bench Press(W) & Upper  & 56.7 & 75   & 62.5 & Abduct arm\\
    Deep Squat(C)  & Lower  & 83.3 & 86.7 & 92.9 & Knee initiation\\
    Deep Squat(W)  & Lower  & 42.2 & 73.3 & 86.7 & Abduct thighs (slow)\\
    Pull-ups(W)    & Upper  & 71.6 & 79.9 & 89.9 & Lats control\\
    $CO_{2}$ Production(W) & Upper & 87.9 & 94.2 & / & Eyes looking straight\\
    Baduanjin(W) & Both & 75.3 & 67.3 & 88.5 & Abduct arm \\
    \bottomrule
  \end{tabular}
  \label{Tbl1}
\end{table}

We then put all of our video triplets into LitePose to detect joint coordinates for pose assessment. The predicted results(joints, center, image id) is constructed into a dictionary with keys in filenames of each frame. The model then normalizes the data and identifies key joints. For example, in bench press, the targeted joints would be shoulders and elbows. As a result, pose assessment examine the given video with the groundtruth exercise video only through shoulders and elbows.

\subsection{Action Assessment}
We create an action quality assessment dataset for the training of our spatial-temporal transformer model. There are 5 professional teachers and 5 students performing 5 different exercises in fitness, rehabilitation and experiment settings. Videos of each subject were recorded in the same front view with diverse camera angle and distances to enhance the robustness of our model on real world settings. Our dataset contains 30 thousand video frames with groundtruth annotation labelled by professional teachers. They provide accurate feedback on each video and we categorize them into two categories: joint position/angle based and pace based. For each exercise video, we label them with three different metrics: joint angle, joint alignment and pace. Joint alignment is calculated by Alg \ref{Alg1}. By providing these additional annotations, the transformer model can learn to attend to the specific joint characteristics and understand the relationships between the annotated features and the overall quality of the exercise performance.

\subsection{Visualized Feedback}
Precise results in pose assessment help pose correction. The corresponding mistake for each exercise is shown in Table 1. The score of joints assessment indicates how well the exercise is performed and thus gives out the visual aid. Visualization is achieved by reconstructing the correct joint position back into every key frame. The result is given in arrows pointing out how the joints should move to improve.

\section*{Conclusion}
Movement assessment and correction provides valuable information about students' performance in education activities. In this paper, we proposed our spatial-temporal transformer based framework for pose assessment and correction. It adopted Human Pose Estimation, Skeleton Tracking, Pose matching, Action Quality Assessment and Pose Correction. We achieved corrective results for the performance of different excises in the form of scores. We also had corrective visual aids as feedback. The STTF framework demonstrates its efficacy in human pose assessment and correction in education scenarios. The framework's ability to capture spatial and temporal dependencies in human poses contributes to accurate assessment and effective correction of students' movements, enhancing the learning experience and facilitating  skill performance. The STTF represents a significant advancement in the field of AI aided education and holds promising potential for improving teaching and learning practices. However, there were also some limitations. Firstly, our framework hasn't been evaluated on a larger number of samples to further testify the reliability. In addition, more professional information for pose correction is needed to generate more accurate solutions for students to adjust. 
\begin{figure}
  \centering
  \includegraphics[width=1.0\linewidth]{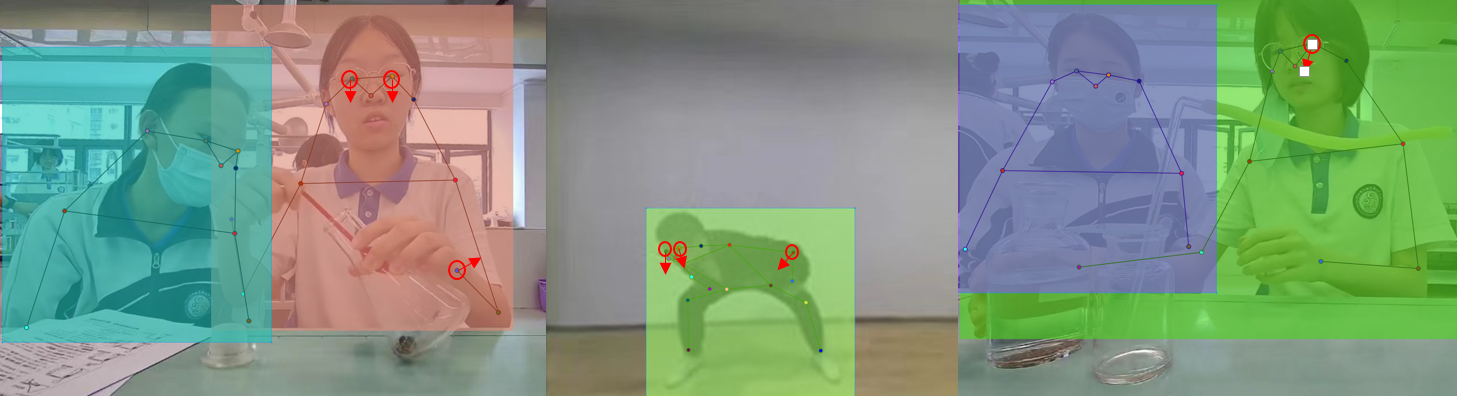}
  \caption{Visual Aid}
  \label{fig6}
\end{figure}

\bibliographystyle{IEEEtran}
\bibliography{references}
\end{document}